\pdfoutput=1

\documentclass[11pt]{article}

\usepackage[final]{acl}

\usepackage{times}
\usepackage{latexsym}
\usepackage{amsmath, amssymb}
\usepackage[T1]{fontenc}

\usepackage[utf8]{inputenc}

\usepackage{microtype}

\usepackage{inconsolata}
\usepackage{graphicx}
\usepackage{booktabs}
\usepackage{enumitem}
\usepackage{multirow}
\usepackage{caption}
\usepackage{subcaption}
\usepackage[dvipsnames]{xcolor}

\title{\Algname{}: Mitigating Confidence Mis-calibration for \\ Large Language Models}

\def\Algname{Self-Ensemble}

\def\Algnameabbr{\texttt{Self-Ensemble}}

\author{
 \textbf{Zicheng Xu\textsuperscript{1,3}\thanks{Equal contribution, ordered by rolling dices.}},
 \textbf{Guanchu Wang\textsuperscript{2}\footnotemark[1]${}^\dagger$},
 \textbf{Guangyao Zheng\textsuperscript{3}},
 \textbf{Yu-Neng Chuang\textsuperscript{1}},
\\
 \textbf{Alexander Szalay\textsuperscript{3}},
 \textbf{Xia Hu\textsuperscript{1}},
 \textbf{Vladimir Braverman\textsuperscript{3}\thanks{Correspondence to: Vladimir Braverman \href{mailto:vova@cs.jhu.edu}{vova@cs.jhu} \\ \href{mailto:vova@cs.jhu.edu}{.edu} and Guanchu Wang \href{mailto:gwang16@charlotte.edu}{gwang16@charlotte.edu}}} 
\\
\\
 \textsuperscript{1}Rice University,
 \textsuperscript{2}University of North Carolina at Charlotte,
 \textsuperscript{3}Johns Hopkins University
}

\begin{document}
\maketitle

\begin{abstract}

Although Large Language Models~(LLMs) perform well in general fields, they exhibit a \textit{\textbf{confidence mis-calibration}} problem on multi-choice question-answering~(MCQA), particularly as the number of answer choices increases.
Specifically, on MCQA with many choices, LLMs suffer from under-confidence in correct predictions and over-confidence in incorrect ones, leading to a substantially degraded performance.
To solve this problem, we propose \Algnameabbr{} in this work.
Our method splits the choices into several groups and ensembles LLM predictions across these groups to reach a final decision.
The advantage of \Algnameabbr{} is its plug-and-play nature, where it can be integrated into existing LLM architecture based on a designed attention mask and positional encoding, without requiring labeled datasets for parameter tuning.
Experimental results on three LLMs and datasets demonstrate that \Algnameabbr{} comprehensively addresses the confidence mis-calibration problem of LLMs, outperforming standard inference as well as baseline methods.
The source code is available at \url{https://github.com/ZichengXu/Self-Ensemble}.

\end{abstract}

\section{Introduction}

Large Language Models~(LLMs) have exhibited remarkable performance in processing natural language information, such as the LLaMA~\cite{touvron2023llama}, Mistral~\cite{jiang2024mixtral}, and Deepseek~\cite{guo2025deepseek}.
Among various NLP tasks, multi-choice question-answering (MCQA) stands out as a standard and challenging benchmark to evaluate the reliability and reasoning ability of LLMs~\cite{wang2024assessing}. 
It can significantly reduce the hallucinations by limiting the answer to a predefined set of choice aligned with human knowledge~\cite{anjum2024halo, neeley2025survey}.

While advanced LLMs perform well on standard MCQA benchmarks, such as MMLU~\cite{hendryckstest2021}, MathQA~\cite{amini-etal-2019-mathqa}, and GSM-8K~\cite{cobbe2021gsm8k}, they still face challenges when it comes to numerous choices that are closely related~\cite{wang2024adaleval}.
In particular, we identify a critical problem: LLMs suffer from \textbf{\emph{confidence mis-calibration}} on MCQA tasks based on a comprehensive benchmark analysis. 
Specifically, LLM's confidence in the correct choice tends to degrade, while its confidence in incorrect choices increases as the number of choices grows.
This problem grows even more pronounced in the presence of additional noisy or partially relevant choices, thereby increasing the unreliable generations and erroneous predictions.

\begin{figure}
    \centering
    \captionsetup{belowskip=-12pt}
    \includegraphics[width=\linewidth]{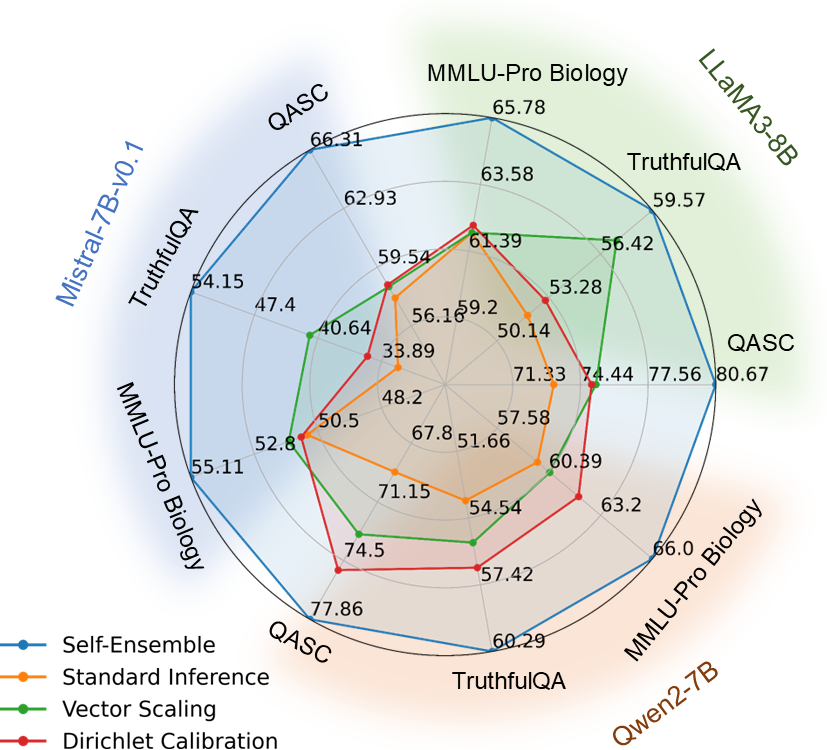}
    \caption{\Algnameabbr{}'s comprehensive performance on the QASC, TruthfulQA, and MMLU-Pro Biology datasets compared with baseline methods.}
    \label{fig:radar}
\end{figure}

\begin{figure*}[ht]
  \centering
  \begin{subfigure}[t]{0.32\textwidth}
    \centering
    \includegraphics[width=\textwidth]{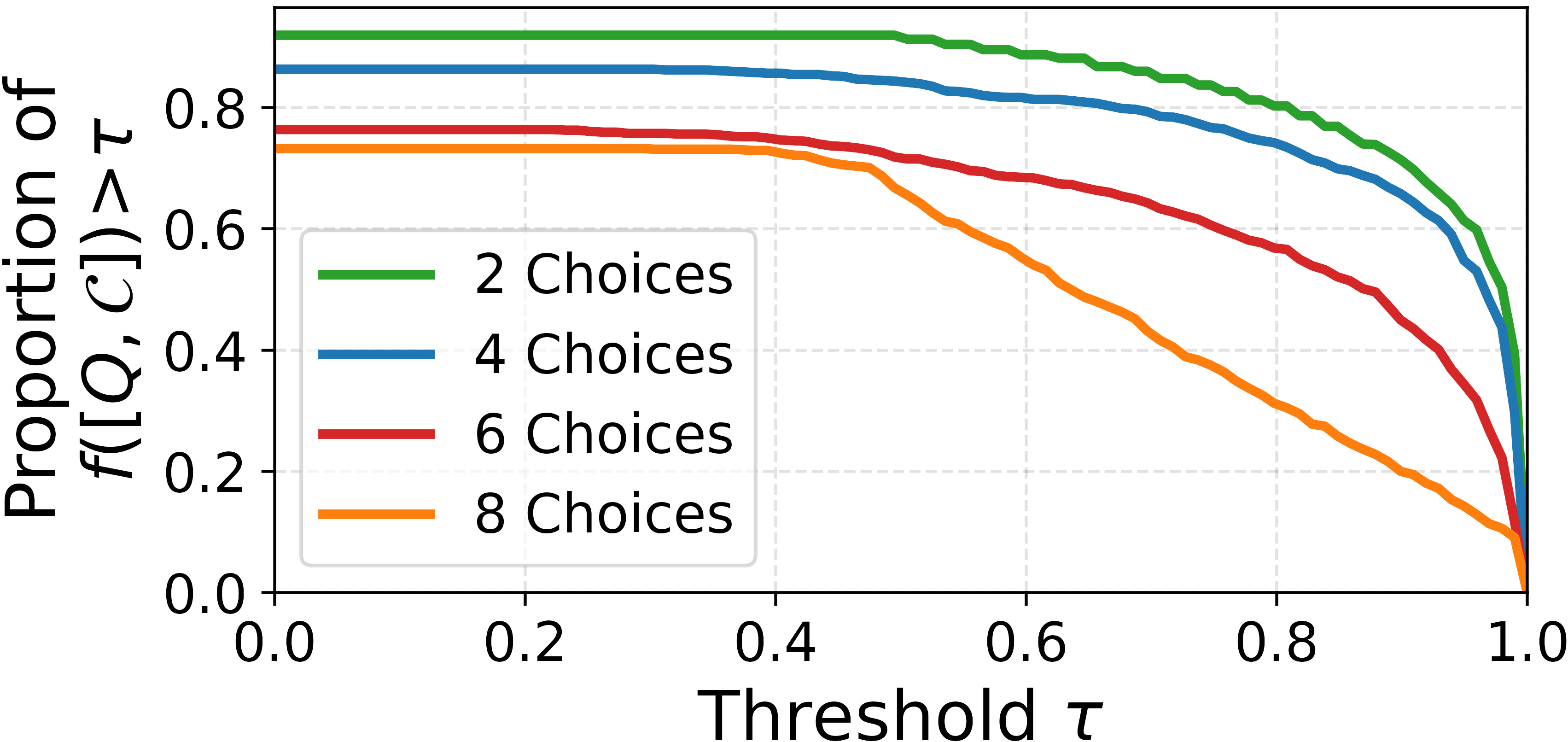}
    \caption{LLaMA-3-8B (Correct)}
    \label{fig:llm_confidence_llama_correct}
  \end{subfigure}\hfill
  \begin{subfigure}[t]{0.32\textwidth}
    \centering
    \includegraphics[width=\textwidth]{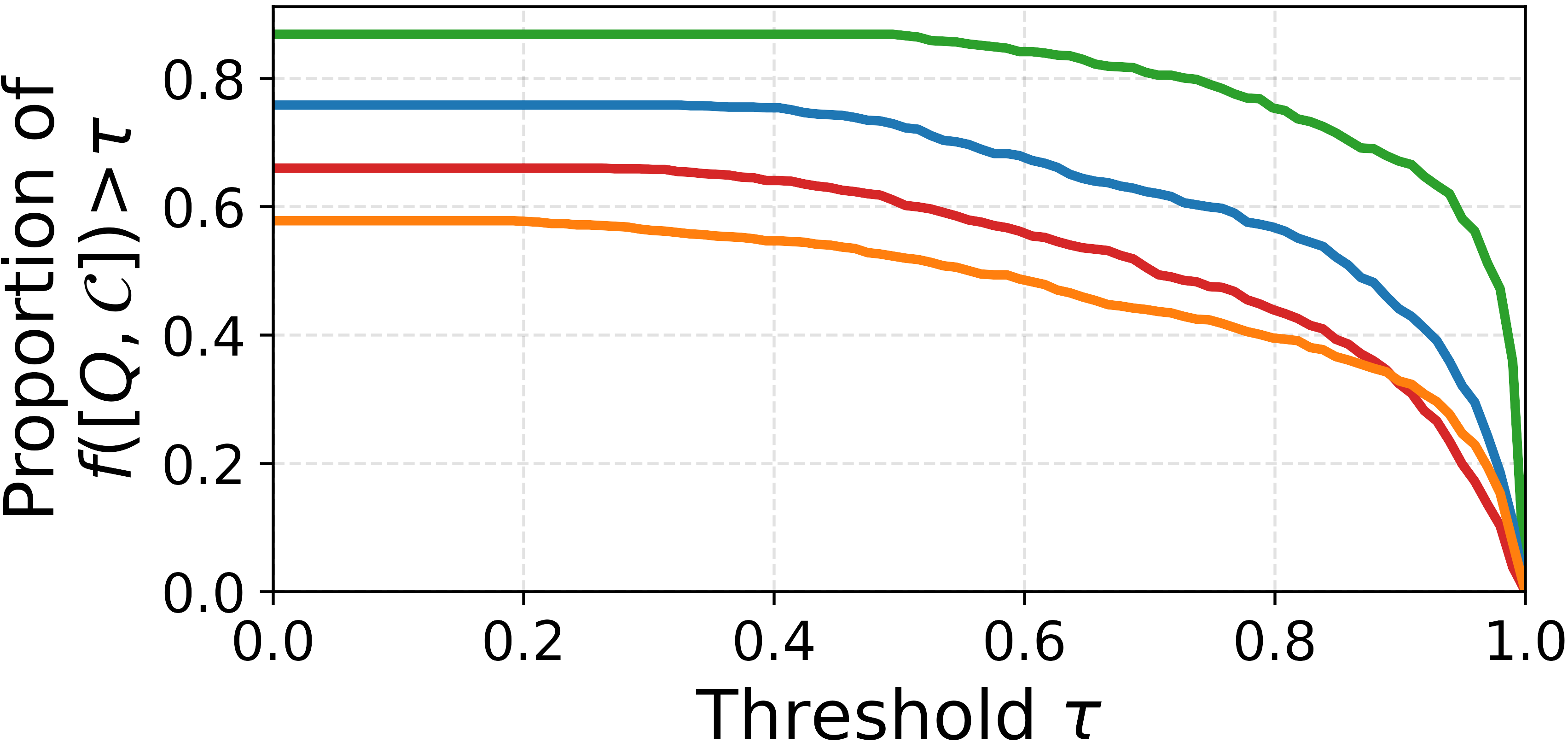}
    \caption{Mistral-7B-v0.1 (Correct)}
    \label{fig:llm_confidence_mistral_correct}
  \end{subfigure}\hfill
  \begin{subfigure}[t]{0.32\textwidth}
    \centering
    \includegraphics[width=\textwidth]{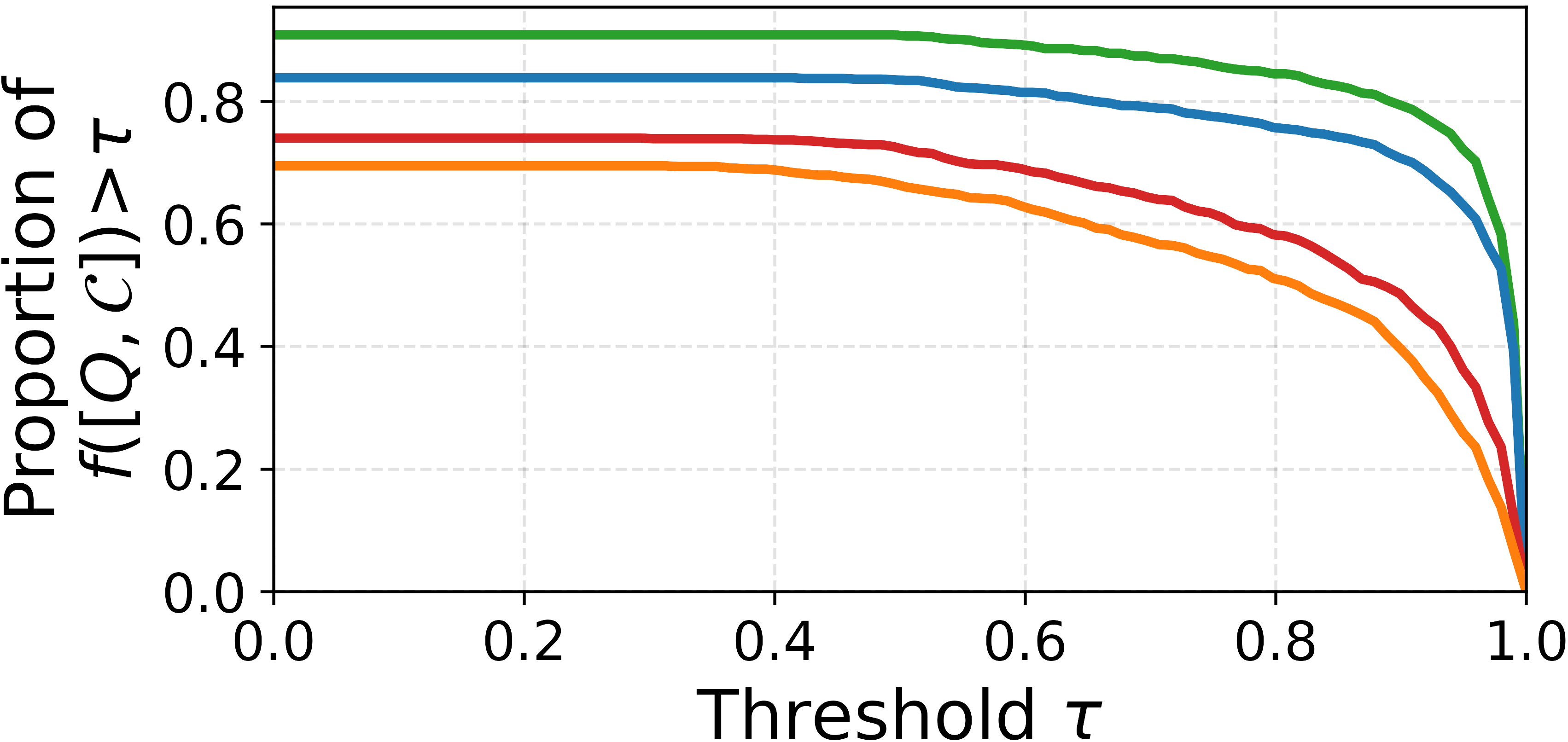}
    \caption{Qwen2-7B (Correct)}
    \label{fig:llm_confidence_qwen_correct}
  \end{subfigure}

  \begin{subfigure}[t]{0.32\textwidth}
    \centering
    \includegraphics[width=\textwidth]{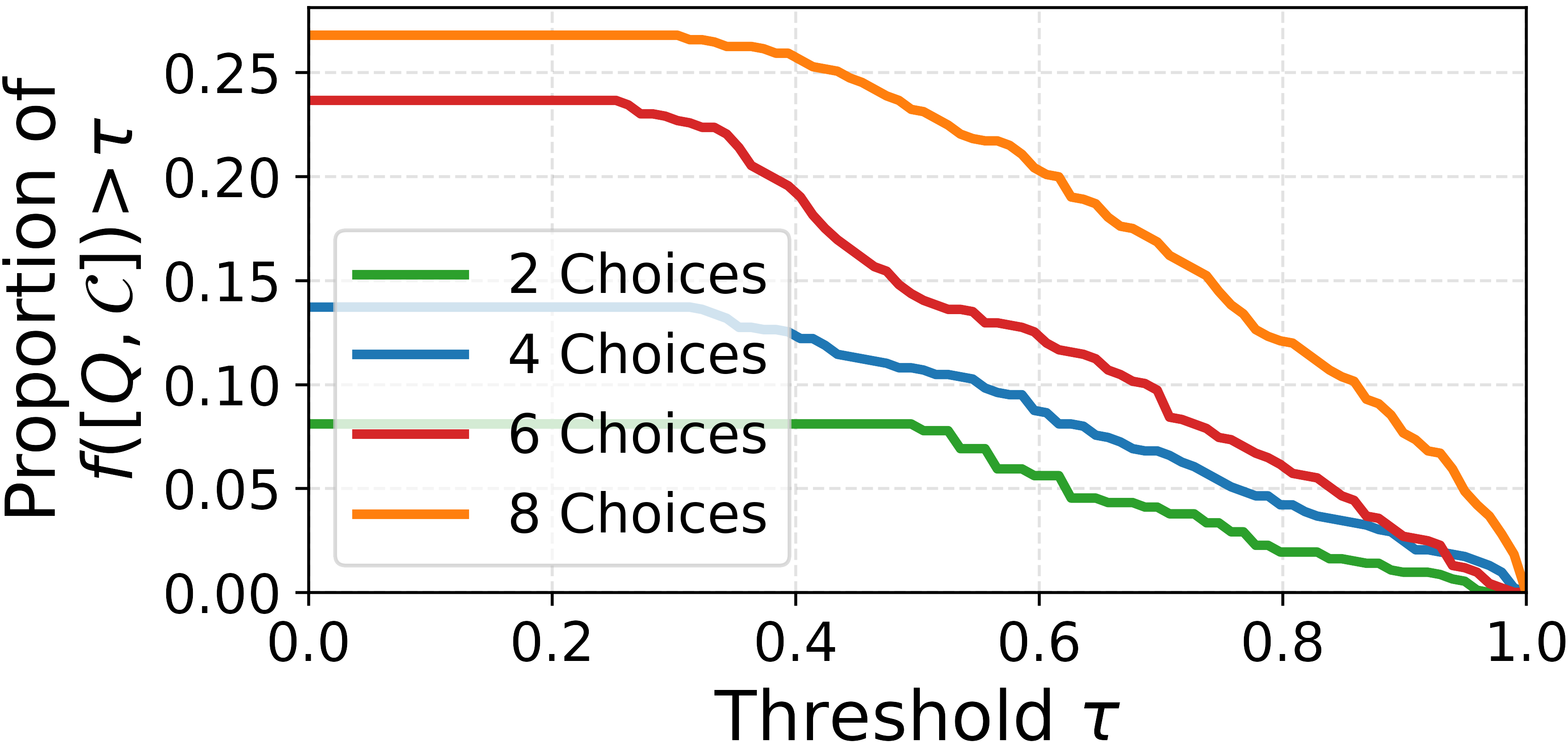}
    \caption{LLaMA-3-8B (Incorrect)}
    \label{fig:llm_confidence_llama_incorrect}
  \end{subfigure}\hfill
  \begin{subfigure}[t]{0.32\textwidth}
    \centering
    \includegraphics[width=\textwidth]{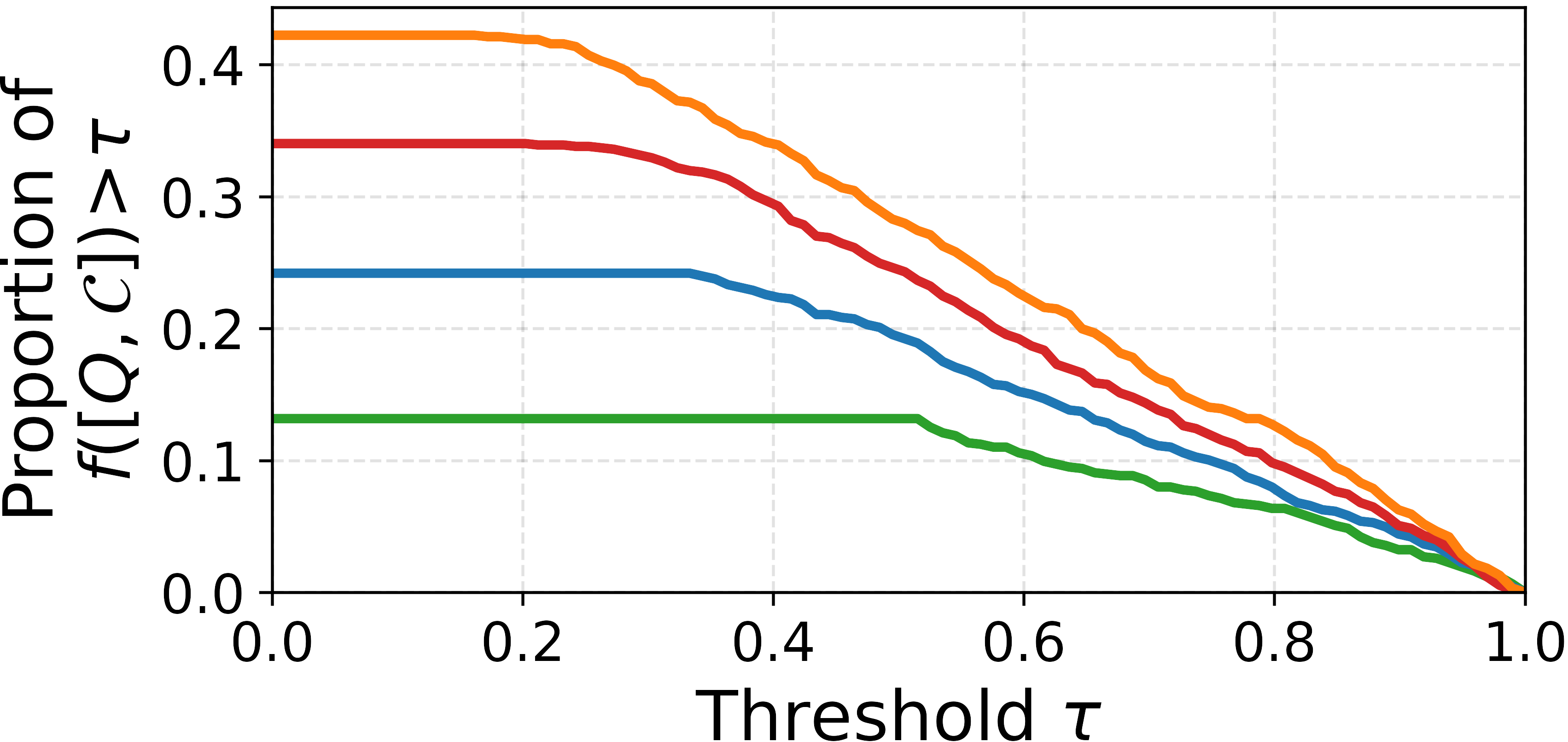}
    \caption{Mistral-7B-v0.1 (Incorrect)}
    \label{fig:llm_confidence_mistral_incorrect}
  \end{subfigure}\hfill
  \begin{subfigure}[t]{0.32\textwidth}
    \centering
    \includegraphics[width=\textwidth]{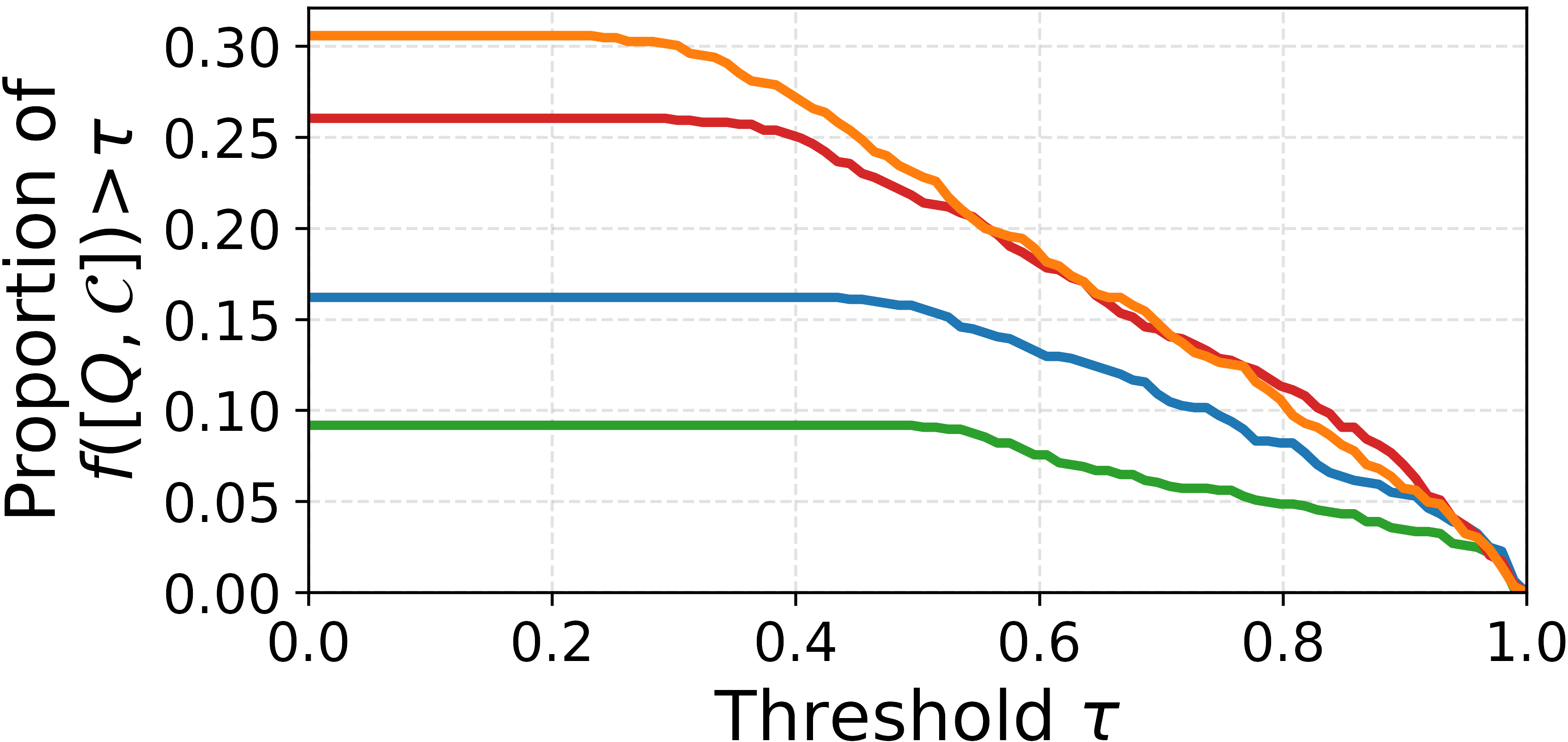}
    \caption{Qwen2-7B (Incorrect)}
    \label{fig:llm_confidence_qwen_incorrect}
  \end{subfigure}

\vspace{-3mm}
  \caption{Proportion of model prediction probability exceeding a threshold on the QASC dataset, for each model under both correct- and incorrect-answer conditions.}
  \label{fig:llm_confidence_all}
\vspace{-4mm}
\end{figure*}

To calibrate LLM predictions on MCQA, existing work focuses on post-processing LLM predictions to better align confidence with correctness.
One common method is the vector scaling~\cite{guo2017calibration}.
It adjusts the model outputs by applying a learned scaling vector and bias vector to the logits before applying the softmax function, which can smooth or sharpen the probabilities to better align with correctness.
Another approach is Dirichlet Calibration~\cite{zong2024dirichlet}. It generalizes LLM's prediction probability on each choice using a Dirichlet distribution to maximize the likelihood function on a validation set.
However, a major limitation of these methods is their reliance on a labeled validation dataset to optimize calibration parameters. 
In practical scenarios, collecting high-quality validation data can be both labor-intensive and costly, which limits these techniques.





To overcome this limitation, we introduce \Algnameabbr{} for LLM calibration.
Unlike existing work, \Algnameabbr{} calibrates LLM predictions without relying on a labeled validation dataset for parameter tuning.
Specifically, \Algnameabbr{} mitigates the confidence mis-calibration through a divide-and-conquer process: it splits the choices into several groups of choice subsets; then ensembles LLM predictions on each group to achieve the final decision.
To plug into existing LLM architecture, we design attention masking and positional re-encoding mechanisms that enable the divide-and-conquer process to be executed internally during inference.
Figure~\ref{fig:radar} shows the comprehensive performance of \Algnameabbr{} on three datasets and LLMs, where our method shows significant improvement over standard inference and baseline methods by addressing the confidence mis-calibration.
To summarize, the contributions of this work are as follows:
\begin{itemize}[
    leftmargin=4mm,
    topsep=1pt,   
    itemsep=1pt,    
    parsep=0pt,     
    partopsep=0pt   
]

    \item \textbf{Confidence Mis-calibration.} We identify the confidence mis-calibration problem of LLMs in MCQA task, where LLMs have under-confident problems on correct answers and over-confident problems on incorrect answers. 
    

    \item \textbf{Self-Ensemble.} We introduce \Algnameabbr{} to calibrate LLM confidence in MCQA task. We design attention masking and positional re-encoding to plug \Algnameabbr{} into existing LLM architectures.

    \item \textbf{Evaluation.} Experimental results demonstrate that \Algnameabbr{} can comprehensively mitigate the under-confident problems on correct answers and over-confident problems on incorrect answers.
    It can effectively improve accuracy on benchmark datasets by 8\% on average.
    
\end{itemize}

\vspace{-2mm}
\section{Preliminary}
\vspace{-2mm}
\subsection{Notations}
We consider an LLM $f$ on MCQA tasks in this work.
We define an MCQA task with $K$ answer choices as \textit{{$K$-choice QA}}, where the set of options is denoted as $\mathcal{C} \!=\! \{ op_1, op_2, \cdots, op_K \}$.
In this work, we refer to the setting with $K \!\leq\! 4$ as \textit{\textbf{few-choice QA}}, and with $K \!\geq\! 8$ as \textit{\textbf{many-choice QA}}.
Our goal is to solve the degradation of LLM's performance and confidence on the task of many-choice QA.

\vspace{-2mm}
\subsection{LLMs for MCQA}

The two most effective ways for LLMs doing MCQA are \emph{verbal generation} and \emph{token probability}.
In the verbal generation approach, the LLM generates the final answer as text outputs, corresponding to one of the choices. 
However, a key drawback is that it may fail to follow the expected output format, leading to answer extraction errors.
In contrast, the token probability approach directly computes the probability of each choice token at the model's final layer, selecting the answer with the highest probability. 
It offers a more reliable evaluation result and often reflects a model’s best achievable performance.
In this work, we consider the token probability approach for MCQA.


\begin{figure}[t]
  \captionsetup{belowskip=-4pt}
  \includegraphics[width=\columnwidth]{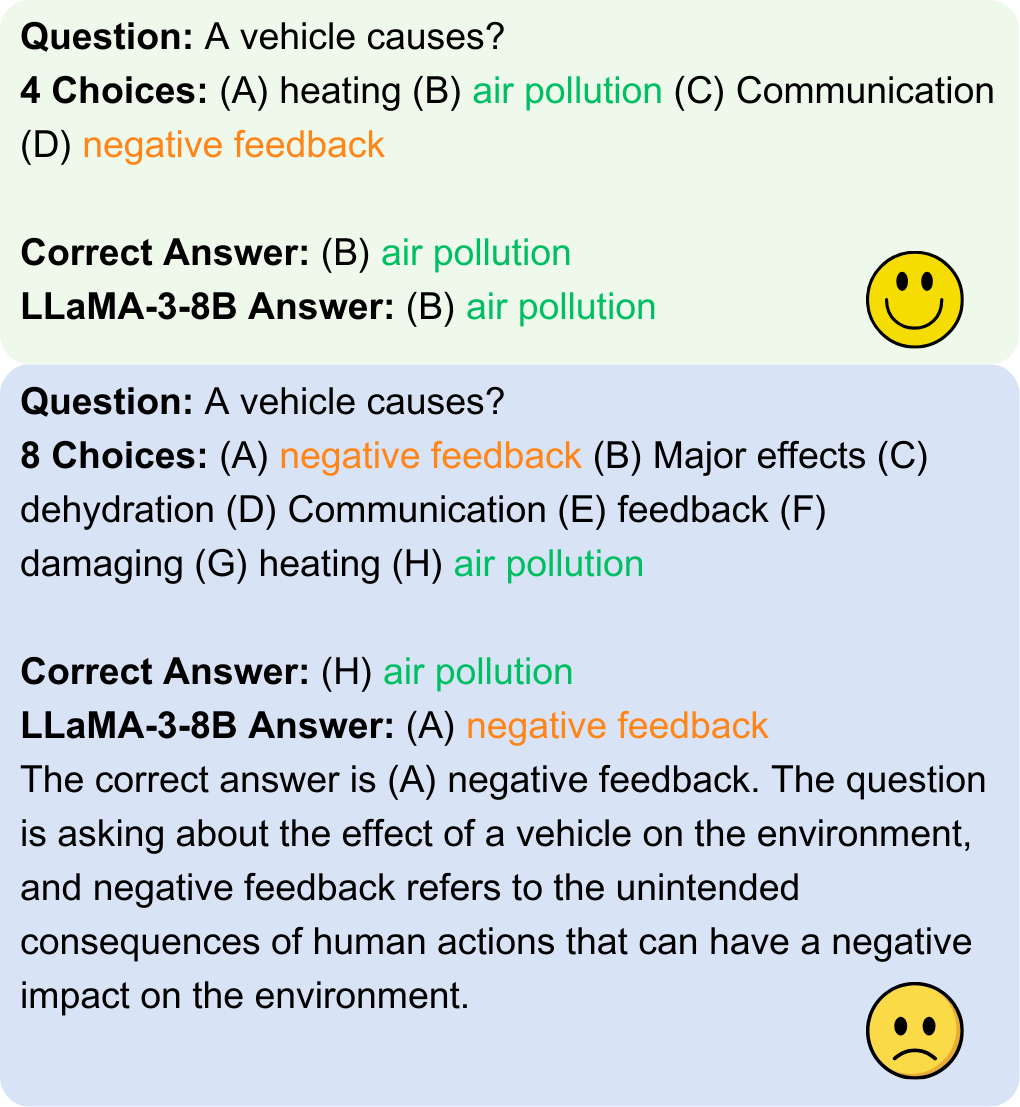}
  \caption{LLMs ignore the correct choice and pick the incorrect one in the many-choice setting.}
  \label{fig:4choices_8choices}
\end{figure}

\begin{table}[t]
    \centering
    \captionsetup{belowskip=-10pt}
    \resizebox{0.5\textwidth}{!}{
    \begin{tabular}{l|cccc}
        \toprule
         \textbf{Model} & \textbf{2-choice} & \textbf{4-choice} & \textbf{6-choice} & \textbf{8-choice} \\ 
         \midrule
         LLaMA-3-8B & \textcolor{OliveGreen}{91.90} & 86.29 & 76.35 & \textcolor{red}{73.22} \\
         Mistral-7B-v0.1 & \textcolor{OliveGreen}{86.83} & 75.81 & 65.98 & \textcolor{red}{57.78} \\
         Qwen2-7B & \textcolor{OliveGreen}{90.82} & 83.80 & 73.97 & \textcolor{red}{69.44} \\ 
         Average  & \textcolor{OliveGreen}{89.85} & 81.97 & 72.10 & \textcolor{red}{66.81} \\ \bottomrule
    \end{tabular}
    }
    \caption{Accuracy of LLMs on the QASC dataset with different choice numbers.}
    \label{tab:confidence_mis-calibration}
\end{table}

\begin{figure*}
    \centering
    \captionsetup{belowskip=-8pt}
    \includegraphics[width=1.0\linewidth]{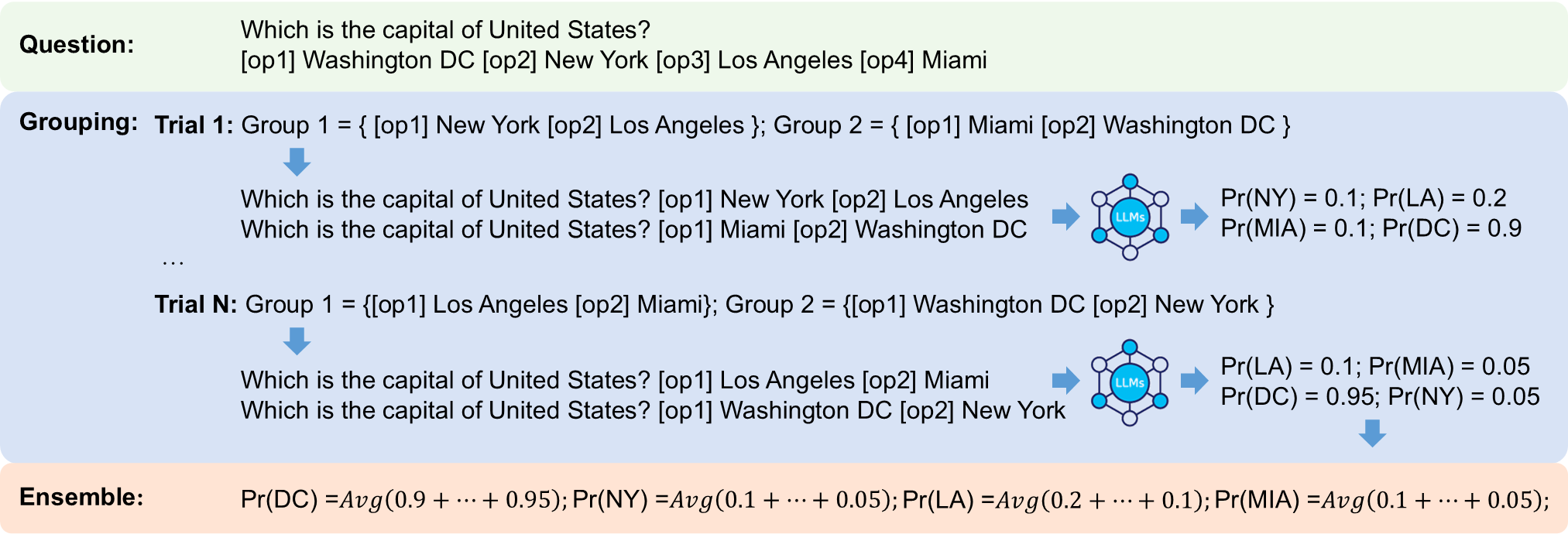}
    \caption{Example of \Algnameabbr{} process on 4-choice QA.}
    \label{fig:selfensemble-example}
\end{figure*}

\section{Confidence Mis-calibration and \Algnameabbr{}}

We demonstrate the \textbf{\textit{confidence mis-calibration}} for LLMs and introduce our \Algnameabbr{} method to overcome this problem.

\subsection{Mis-calibration on Many-choice QA}


We give Figure~\ref{fig:llm_confidence_all} to demonstrate the confidence mis-calibration of LLMs on many-choice QA, where the experiment is on the QASC datasets~\cite{khot2020qasc}.
Specifically, Figure~\ref{fig:llm_confidence_all} shows the proportion of $f([Q, \mathcal{C}]) \!>\! \tau$ versus $\tau$ for $0 \!<\! \tau \!<\! 1$, where $f([Q, \mathcal{C}]) \!>\! \tau$ represents LLM's predicted probability on the correct or incorrect choice exceeds a threshold $\tau$.
A high value of $f([Q, \mathcal{C}]) \!>\! \tau$ proportion across $0 \!\leq\! \tau \!\leq\! 1$ indicates strong model confidence.
To show the confidence mis-calibration, we compare the $f([Q, \mathcal{C}]) \!>\! \tau$ proportion across different choice settings with $K=2, 4, 6, 8$ in Figure~\ref{fig:llm_confidence_all}.
On the correctly answered questions, the LLM achieves the highest $f([Q, \mathcal{C}]) \!>\! \tau$ proportion when $K \!=\! 2$, showing that the model becomes under-confident in many-choice scenarios.
In contrast, for the incorrectly answered questions, the LLM shows the highest $f([Q, \mathcal{C}]) \!>\! \tau$ proportion when $K \!=\! 8$, indicating an over-confidence issue in many-choice scenarios.

The problem of confidence mis-calibration causes a degradation of LLM performance on many-choice QA.
As shown in Figure~\ref{fig:4choices_8choices}, for the LLaMA-3-8B model, it can originally solve the 4-choice QA.
However, when the same question is extended to 8 choices, while keeping the correct answer unchanged, the model selects an incorrect option.
This failure is due to a decrease in confidence in the correct answer and an increase in confidence in incorrect options.
We show a comprehensive result in Table~\ref{tab:confidence_mis-calibration}, where each column shares the same question context but varies in the number of answer choices.
A many-choice question contains all the options from the corresponding few-choice versions.
For example, each question in the 8-choice task includes all the choices from the 6-choice task.
As shown in Table~\ref{tab:confidence_mis-calibration}, LLMs generally perform worse on many-choice QA than few-choice QA, 
This indicates the confidence mis-calibration for LLM on many-choice QA.



\subsection{LLM Inference with \Algnameabbr{}}
\label{sec:self-ensumble}

We propose \Algnameabbr{} to address the confidence mis-calibration on many-choice QA.
The intuition of \Algnameabbr{} is to divide a $K$-choice (many-choice) QA into multiple $m$-choice (few-choice) QA, where $m \!\ll\! K$.
Then, it collects LLM's answer probability in each few-choice case, and estimates the expected probability of each choice for the final decision.
Specifically, \Algnameabbr{} has three steps as follows:

\paragraph{Step 1.} \!\!\!\!\! Given the $K$ choices $\mathcal{C} \!=\! \{ {op}_1, \cdots \!, {op}_K \}$, \Algnameabbr{} randomly splits them into $\lceil \frac{K}{m} \rceil$ groups with seed $s$, which is given by $\mathcal{G}_s(\mathcal{C}) = \{ {G}_1, \cdots, {G}_{\lceil \frac{K}{m} \rceil} \}$.
Each group ${G}_i$ has $m$ choices, while the last group may have less than $m$ choices if $K\!\!\mod{m}\!\neq\! 0 $.
Different groups take random subsets of $\mathcal{C}$ without overlaps, i.e. ${G}_i \cap {G}_j = \varnothing$ for $i \neq j$, and ${G}_1 \cup \cdots \cup {G}_{\lceil \frac{K}{m} \rceil} = \mathcal{C}$.

\paragraph{Step 2.} For each group ${G}_j = \{ \widetilde{op}_1, \cdots, \widetilde{op}_{\tilde{m}} \}$, \Algnameabbr{} collects the probability of each choice within this group from LLMs as follows:
\begin{equation}
\label{eq:self-ensemble-step2}
   Pr(\widetilde{op}_1), \cdots, Pr(\widetilde{op}_{\tilde{m}})  = f([Q, {G}_j])
\end{equation}


\paragraph{Step 3.} To estimate the final probability of each choice $op_i$, \Algnameabbr{} averages all the probabilities of choice $op_i$ as follows:
\vspace{-5pt}
\begin{align}
\label{eq:self-ensemble-step3}
    \!\!\!\! {Pr}(op_i) \!=\! \mathbb{E}_{s \sim U(\mathbb{N}), {G}_j \sim \mathcal{G}_s(\mathcal{C}), \atop \widetilde{op} \sim {G}_j} \!\! \big[ \mathbb{I}(op_i \!=\! \widetilde{op}) Pr(\widetilde{op}) \big] \!\!\!\!\!
\vspace{-2mm}
\end{align}
where $s \!\sim\! U(\mathbb{N}), {G}_j \!\sim\! \mathcal{G}_s(\mathcal{C})$ represents randomly splitting the choice set using different integer random seed $s$; 
and $\mathbb{I}(op_i \!=\! \widetilde{op}) \!=\! 1$ if $op_i \!=\! \widetilde{op}$; otherwise it takes $0$.
The intuition of Equation~(\ref{eq:self-ensemble-step3}) is to estimate the expected probability of each choice by averaging over different group partitions.


An example of \Algnameabbr{} is given in Figure~\ref{fig:selfensemble-example}.
Specifically, in each trial, \Algnameabbr{} simplifies the given 4-choice QA into 2-choice QAs by randomly grouping the choices, and achieves the choice probability within each group from LLMs. 
After $N$ trials, it estimates the expected probability of each choice by averaging different trials.


\begin{figure*}
    \centering
    \captionsetup{belowskip=-8pt}
    \includegraphics[width=1.0\linewidth]{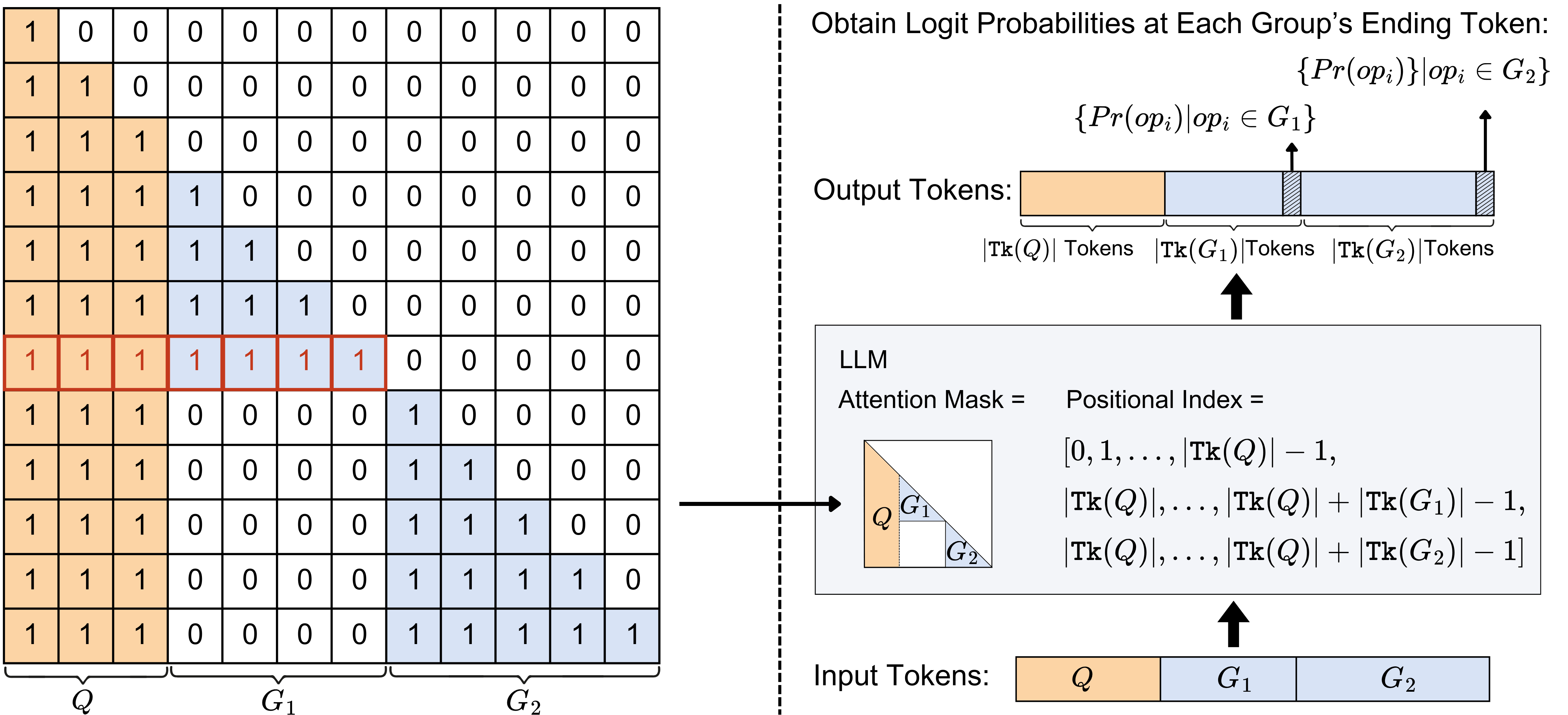}
    \caption{Plug-in \Algnameabbr{}: by incorporating the attention mask and positional re-encoding, LLMs can achieve the ensembled results in a single forward pass.}
    \label{fig:plugin_selfensemble-example}
\end{figure*}

\section{Plugging \Algnameabbr{} into LLMs}
\label{sec:plugin_self_ensemble}


We propose integrating \Algnameabbr{} into existing LLM architectures to enable intrinsic self-ensemble processing during inference. 
As illustrated in Equations~(\ref{eq:self-ensemble-step2}) and (\ref{eq:self-ensemble-step3}), \Algnameabbr{} requires multiple forward passes through the LLM, which are $f([Q, {G}_1])$, $f([Q, {G}_2])$, $f([Q, {G}_3]), \cdots$.
For efficient self-ensemble processing, we design specialized attention masks and positional encoding, such that these multiple calls of LLM can be executed in a single forward pass over the concatenated sequence $f([Q, {G}_1, {G}_2, {G}_3, \cdots])$.
Here, each group ${G}_j \sim \mathcal{G}_s(\mathcal{C}), s \sim \mathbb{N}$ is sampled from the choice set following Section \ref{sec:self-ensumble} Step 1; and $[Q, {G}_1, {G}_2, {G}_3, \cdots]$ denotes combining the question and different groups of choice into a sequence of tokens as prompts for LLMs.

\subsection{Attention Masking}
\label{sec:attention-mask}






The standard auto-regressive attention mask follows $\mathbb{I}(j \leq i)$, ensuring that each token attends to all preceding tokens in the sequence. 
However, following \Algnameabbr{}, the LLM should restrict its attention to a single group of choices during each trial, ignoring the influence of other groups.
Therefore, for the concatenated input sequence with the question and all choice groups $[Q, {G}_1, {G}_2, {G}_3, \cdots]$, \textit{\textbf{when processing a certain choice group $G_i$, the LLM should not have attention to other groups $G_j$, $j \neq i$}}.

To implement this constraint, \Algnameabbr{} designs a custom attention mask to block the cross-group attention. 
As shown in the left-hand side of Figure~\ref{fig:plugin_selfensemble-example}, a token within group $G_1$ (highlighted in red) attends only to the question and other tokens within $G_1$. 
This ensures that the model focuses solely on the current group when generating the answer.
In general, let $\texttt{Tk}_i$ denote a certain token at position $i$. 
Its attention mask to a previous token $\texttt{Tk}_j$, $j \leq i$ is given by
\begin{equation}
  \label{eq:mask_complete}
  \!\!\!\!\! M_{i,j} \!=\!
  \begin{cases}
    1, 
    & \!\!\!\!\!\!\!\!\!\!\!\!\!\!\!\!\! \text{if } \texttt{Tk}_j \in \texttt{Tk}(Q), \\
    \prod\limits_{n=1}^N \mathbb{I}(\texttt{Tk}_i, \texttt{Tk}_j \in \texttt{Tk}(G_n)) 
    & \!\!\!\!\text{otherwise}, \!\!\!\!\!\!
  \end{cases}
\end{equation}
where $\texttt{Tk}(Q)$ denotes the tokens of question $Q$; $\mathbb{I}(\texttt{Tk}_i, \!\texttt{Tk}_j \!\in\! \texttt{Tk}(G_n)) \!=\! 1$ if $\texttt{Tk}_i$ and $\texttt{Tk}_j$ belong to the same group, and $0$ otherwise; and $M_{i,j} = 0$ for $j > i$ as standard attention masks.
This attention mask enforces pairwise independence between choice groups during inference.

\subsection{Positional Re-encoding}
\label{sec:positional-encoding}




The standard auto-regressive positional encoding follows $\texttt{Pos}_i \!=\! i$, where $\texttt{Pos}_i$ denotes the positional index of a token at position $i$.
Following \Algnameabbr{}, the LLM should process each question-group pair $[Q, {G}_j]$ as a continuous input sequence for $1 \!\leq\! j \!\leq\! N$.
However, this continuity is disrupted in the concatenated input sequence $[Q, G_1, G_2, G_3, \cdots]$, because starting from $G_2$, the question $Q$ and a choice group are no longer physically adjacent.
For example, ${G}_1$ lies between $Q$ and ${G}_2$, breaking the positional continuity between $Q$ and $G_2$.
To preserve the relative positional relationships between the question and each choice group, \textit{\textbf{each choice group should be encoded into the relative contextual position to the question}}.
Specifically, the positional indexes of question tokens and choice tokens are given as follows:


\vspace{-2pt}
\paragraph{Question tokens.} A question token $\texttt{Tk}_i \!\in\! \texttt{Tk}(Q)$ should have absolute position $\texttt{Pos}_i = i$.


\vspace{-2pt}
\paragraph{Choice tokens.} A choice token $\texttt{Tk}_i$, if it is with choice group $G_n$, its absolute position should be replaced into a relative position as follows:
\begin{equation}
  \label{eq:pos_choice}
  \texttt{Pos}_i = i - \sum_{j=1}^n |\texttt{Tk}(G_j)|,
\end{equation}
where $|\texttt{Tk}(G_j)|$ indicates the token number of a choice group $G_j$; and $n$ takes maximal value that satisfies $i \geq \sum_{j=1}^n |G_j|$.

Following Equation~(\ref{eq:pos_choice}) to replace the position index of choice tokens, the position index of input sequence $[Q, {G}_1, \cdots, {G}_n]$ is given by
\vspace{-2pt}
\begin{align}
\label{eq:new-position}
    &\Big[\overbrace{0,1, \cdots, |\texttt{Tk}(Q)|\!-\!1}^{\text{Position index of Q}}, \!\!\!\!\!\!\!\!
    \\
    &\overbrace{|\texttt{Tk}(Q)|, \cdots, |\texttt{Tk}(Q)|\!+\!|\texttt{Tk}(G_1)|\!-\!1}^{\text{Position index of $G_1$}}, \cdots,
    \nonumber
    \\
    &\overbrace{|\texttt{Tk}(Q)|, \cdots, |\texttt{Tk}(Q)|\!+\!|\texttt{Tk}(G_n)|\!-\!1}^{\text{Position index of $G_n$}}\Big]
    \nonumber
\end{align}

By combining the attention mask with positional re-encoding, we ensure that processing the concatenated sequence $[Q, G_1, G_2, G_3, \cdots]$ is functionally equivalent to independently processing each question-group pair $[Q, G_1], [Q, G_2], [Q, G_3], \cdots$.
This enables \Algnameabbr{} to seamlessly integrate into existing LLM architectures.






\subsection{Choice Probability}


By using the attention mask and positional re-encoding in Equations~(\ref{eq:mask_complete}) and (\ref{eq:new-position}), respectively, the LLM can simultaneously process all groups of choices in a single forward pass.
As shown in Figure~\ref{fig:plugin_selfensemble-example}, within the output sequence, the ending token of each group captures the choice probabilities  within each choice group.
Formally, let $\boldsymbol{a} = f([Q, {G}_1, {G}_2, {G}_3, \cdots])$ denote the sequence of output logits.
For a certain choice $\widetilde{op}_i \in {G}_j$, the probability of $\widetilde{op}_i$ is estimated by
\begin{equation}
Pr(\widetilde{op}_i)  = \boldsymbol{a}[\texttt{index}_j, \texttt{index}_i],
\end{equation}
where $\texttt{index}_j \!=\! |Q| \!+\! \sum_{k=1}^{j-1} \! |\texttt{Tk}(k)| \!-\! 1$ takes the ending token position of group $G_j$; $\texttt{index}_i$ takes the token ID of $\widetilde{op}_i$ defined by the LLM tokenizer.
The final probability of each choice takes the expected probability $Pr({op}_i) \!=\! \mathbb{E} [ \mathbb{I}({op}_i \!=\! \widetilde{op}_i) Pr(\widetilde{op}_i) ]$.


The advantage of \Algnameabbr{} lies in its seamless integration with existing LLM architectures.
It enables the LLM to achieve the expected probability of each choice in a single forward pass, avoiding separate inference over individual choice groups.
This enables \Algnameabbr{} to behave both accurately and efficiently on MCQA.



\begin{table*}[h]
    \centering
    \captionsetup{belowskip=-10pt}
    \resizebox{\textwidth}{!}{
    \begin{tabular}{l|l|cc|cc|cc|cc}
    \toprule
         & & \multicolumn{2}{c|}{QASC} & \multicolumn{2}{c|}{TruthfulQA} & \multicolumn{2}{c|}{MMLU-Pro Biology} & \multicolumn{2}{c}{Average} \\
    \midrule
         Model & Method & Accuracy & Improve & Accuracy & Improve & Accuracy & Improve & Accuracy & Improve \\
    \midrule
        \multirow{4}{*}{LLaMA-3-8B} & Standard inference         & 73.22 & - & 51.99 & - & 62.00 & - & 62.40 & -\\
        & Vector Scaling      &    75.16 & +1.94   &   57.40  & +5.41  &   62.00 & +0.00 & 64.85 & +2.45    \\
        & Dirichlet Calibration    &   74.96 & +1.74    &  53.07 & +1.06    &   62.24 & +0.24 & 63.42 & +1.02  \\
        & \Algnameabbr{}           & \textbf{80.67} & \textbf{+7.45} & \textbf{59.57} & \textbf{+7.58}& \textbf{65.78} & \textbf{+3.78} & \textbf{68.67} & \textbf{+6.27}\\
        \midrule
        \multirow{4}{*}{Mistral-7B-v0.1} & Standard inference         & 57.78 & - & 32.13 & - & 50.89 & - & 46.93 & - \\
        & Vector Scaling      &   58.42 & +0.64    &  41.52 & +9.39     &   51.56 & +0.67  & 50.50 & +3.57  \\
        & Dirichlet Calibration    &   58.53 & +0.75    &  35.38 & +3.25     &  51.11 & +0.22 & 48.34 &+1.41    \\
        & \Algnameabbr{}           & \textbf{66.31} & \textbf{+8.53} & \textbf{54.15} & \textbf{+22.02} &\textbf{55.11} & \textbf{+4.22} & \textbf{58.52} & \textbf{+11.59} \\
        \midrule
        \multirow{4}{*}{Qwen-2-7B} & Standard inference         & 69.44 & - & 53.79 & - & 59.78 & - & 61.00 & -\\
        & Vector Scaling      &   73.00 & +3.56    &   55.60 & +1.81   &   60.44 & +0.66  & 63.01 & +2.01  \\
        & Dirichlet Calibration    &   75.05 & +5.61    &  56.68 & +2.89    &  62.00 & +2.22  & 64.58 & +3.58     \\
        & \Algnameabbr{}           & \textbf{77.86} & \textbf{+8.42} & \textbf{60.29} & \textbf{+6.50} & \textbf{66.00} & \textbf{+6.22} & \textbf{68.05} & \textbf{+7.05} \\
    \bottomrule
    \end{tabular}
    }
    \caption{Accuracy of \Algnameabbr{} on the QASC, TruthfulQA, MMLU-Pro Biology datasets.}
    \label{tab:exp_acc}
\end{table*}

\section{Experiment}

In this experiment, we conduct experiments to evaluate \Algnameabbr{}, aiming to answer the following research questions: 
\textbf{RQ1:} Can \Algnameabbr{} improve on LLM's accuracy on many-choice QA?
\textbf{RQ2:} Can \Algnameabbr{} mitigate the confidence mis-calibration problems for LLMs?
\textbf{RQ3:} How does \Algnameabbr{} reshape the scaling of model accuracy with respect to parameter size?
\vspace{-2pt}
\subsection{Experimental Setup}

We specify the datasets, LLMs, evaluation metrics, and implementation details.

\paragraph{Models.} We evaluate \Algnameabbr{} using three popular model families: LLaMA-3-8B~\cite{touvron2023llama}, Mistral-7B-v0.1~\cite{jiang2024mixtral}, and Qwen-2-7B~\cite{qwen2}.
We download these models from the Huggingface Transformers library~\cite{wolf2019huggingface}.
\vspace{-2pt}
\paragraph{Dataset.} The evaluation of \Algnameabbr{}~is based on the QASC~\cite{khot2020qasc}, TruthfulQA~\cite{lin2021truthfulqa}, and MMLU-Pro~\cite{wang2024mmlu} datasets. \textbf{QASC:} It is a multi-hop reasoning dataset comprised of 8-choice QA. We evaluate LLM performance on its validation set with 927 questions under a closed-book setting. 
\textbf{TruthfulQA:} It has 817 multiple-choice and short-answer questions across 38 categories, such as health, law, and finance. 
Following many-choice setting, we filter out questions with fewer than five incorrect choices.
As a result, 277 questions have at least one correct choice and five incorrect choices.
\textbf{MMLU-Pro:} Enhancement of the MMLU dataset by selecting questions where state-of-the-art LLMs consistently fail. 
We use the Biology subset for our experiments, which has 450 10-choice questions. 



\paragraph{Baseline Methods.}
\textbf{Standard inference:} We perform standard inference by comparing the probability of each choice token and selecting the option with the highest likelihood.  
\textbf{Vector scaling:} Vector scaling~\cite{guo2017calibration} applies a learnable scale vector $w$ and bias $b$ to the logits before softmax, enabling finer-grained calibration. The parameters are optimized to maximize accuracy on a validation set.
\textbf{Dirichlet calibration:} Dirichlet calibration~\cite{zong2024dirichlet} is a lightweight multiclass calibration method that adjusts the model’s predicted probability to better match observed frequencies on a validation set.  

\paragraph{Implementation Details.}
For QASC's 8-choice QA, \Algnameabbr{} splits each question into a 4-choice QA for 20 trials.
In addition, for the TruthfulQA dataset, it splits the 6-choice QAs into 3-choice QAs for 6 trials.
Moreover, for the MMLU-Pro Biology dataset, it splits the 10-choice QAs into 5-choice QAs for 40 trials.
\Algnameabbr{} does not reply on a validation set for optimizing any parameter or special setting.



\begin{figure*}
    \centering
    \includegraphics[width=1.0\linewidth]{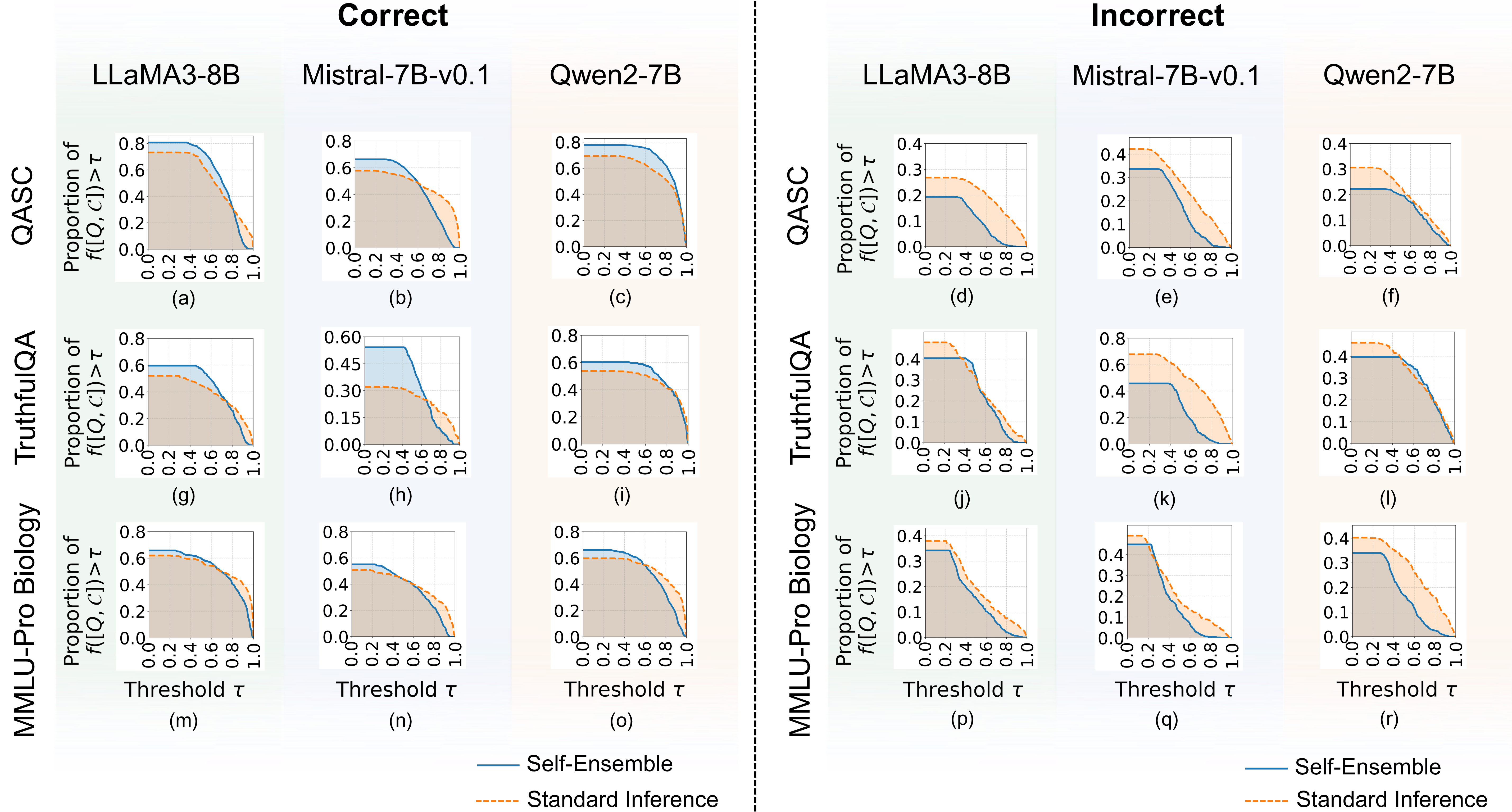}
    \caption{Probability of model confidence exceeding a threshold on the QASC, TruthfulQA, and MMLU-Pro-Biology dataset, for each model under both correct- and incorrect-answer conditions.}
    \label{fig:main_exp_confidence}
\vspace{-3mm}
\end{figure*}

\subsection{Accuracy of MCQA~(RQ1)}

Table~\ref{tab:exp_acc} shows the accuracy~(\%) of \Algnameabbr{}.
These results are compared with baseline methods and standard inference of LLMs.

\paragraph{Accuracy Improvement.}
As shown in Table~\ref{tab:exp_acc}, \Algnameabbr{} consistently outperforms LLM standard inference and baseline methods, demonstrating its potential in enhancing LLM's performance on MCQA.
Moreover, compared with baseline methods, \Algnameabbr{} does not require a validation set for maximizing performance.

\paragraph{Model Agnosticism.}
By integrating with different families of LLMs, \Algnameabbr{} shows consistent performance and improvement, as shown in Table~\ref{tab:exp_acc}.
This generality indicates that \Algnameabbr{} can potentially serve as a versatile enhancement for a range of LLMs in practice.
\vspace{-2pt}
\paragraph{Stable Improvement.}
For datasets of different difficulty levels, standard inference shows less accuracy on TruthfulQA and MMLU-Pro than QASC. 
This implies that the confidence mis-calibration is a general problem on both easy and hard MCQA tasks.
In contrast, \Algnameabbr{} consistently improves performance across all difficulty levels, demonstrating its general effectiveness in mitigating the confidence mis-calibration problem.

\vspace{-2pt}
\subsection{Confidence Calibration by \Algnameabbr{}~(RQ2)}
\vspace{-2pt}

In this section, we show that \Algnameabbr{} mitigates the confidence mis-calibration of LLMs in Figure~\ref{fig:main_exp_confidence}.
Specifically, Figure~\ref{fig:main_exp_confidence} shows the proportion of $f([Q, \mathcal{C}]) \!>\! \tau$ versus $\tau$ for $0 \!<\! \tau \!<\! 1$ on three different LLMs and datasets, where $f$ takes different LLMs; and $f([Q, \mathcal{C}]) \!>\! \tau$ represents LLM's predicted probability exceeds a threshold $\tau$.
A higher value of $f([Q, \mathcal{C}]) \!>\! \tau$ proportion across $0 \!<\! \tau \!<\! 1$ indicates strong model confidence.

In Figure~\ref{fig:main_exp_confidence}'s sub-figures (a)-(c), (g)-(i), (m)-(o), \Algnameabbr{} has a higher confidence level on its correct answers; while in sub-figures~(d)-(f), (j)-(l), (p)-(r), \Algnameabbr{} has a lower confidence level on its incorrect answers. 
This indicates \Algnameabbr{} can effective mitigate the under-confident problems on correct answers and over-confident problems on incorrect answers, improving the reliability of LLMs on MCQA in practice.



\begin{figure}[t]
  \captionsetup{belowskip=-12pt}
  \includegraphics[width=\columnwidth]{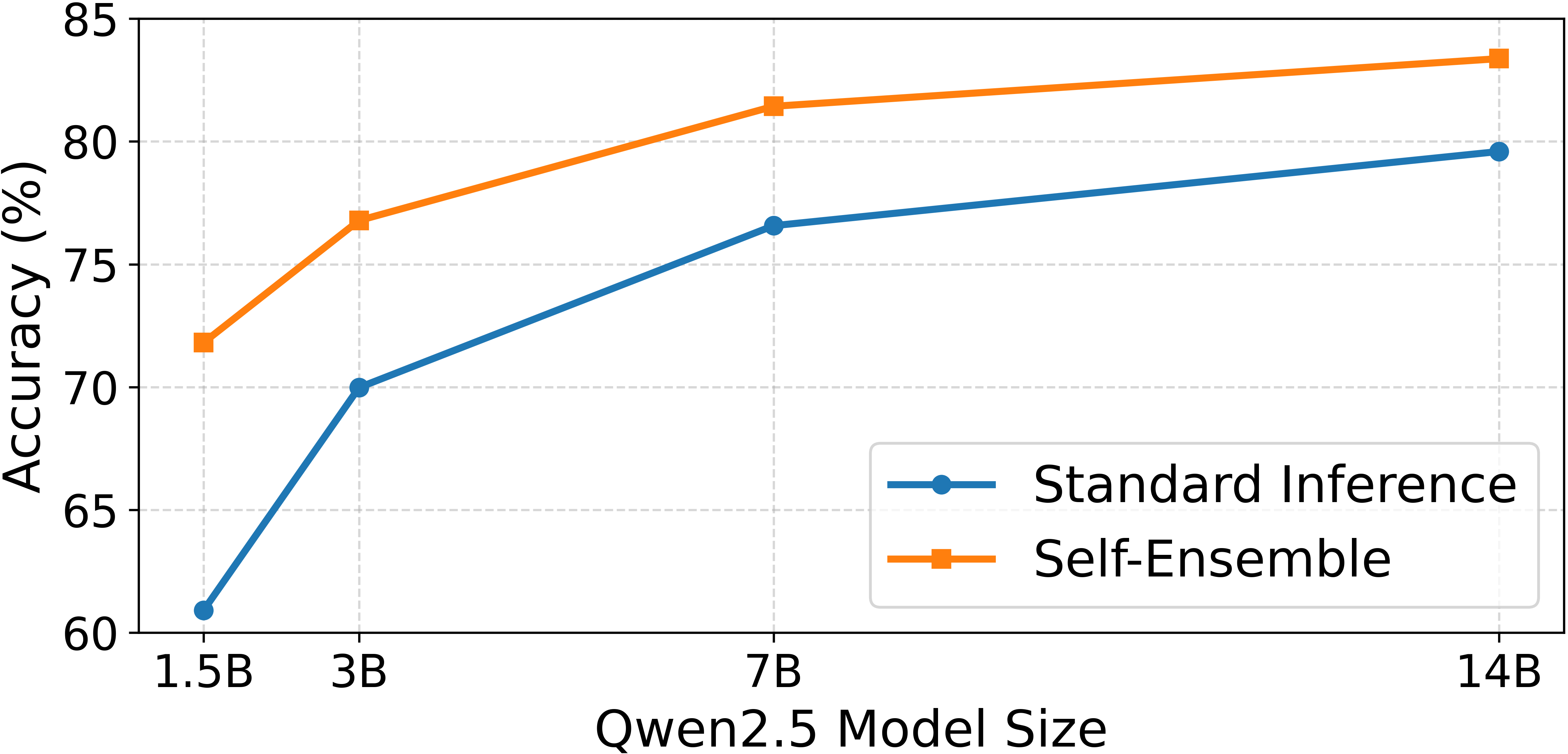}
    \caption{Qwen2.5-1.5B, 3B, 7B, and 14B with standard inference and \Algnameabbr{} on the QASC dataset.}
    \label{fig:scaling_law}
\end{figure}

\vspace{-2pt}
\subsection{Improving Accuracy-to-Parameter Scaling~(RQ3)}

\Algnameabbr{} achieves improved scaling of accuracy with parameter size.
As shown in Figure~\ref{fig:scaling_law}, \Algnameabbr{} consistently yields higher accuracy on the QASC dataset as the model size increases from 1.5B to 14B.
It enables better scalability across model sizes and can be integrated with any LLM without requiring additional data or retraining.
Notably, Qwen2.5-1.5B, 3B, 7B with \Algnameabbr{} outperform larger models using standard inference.
This demonstrates the potential of \Algnameabbr{} to advance the deployment of small language models in real-world scenarios with constrained memory resources.

\subsection{General-purpose LLMs with \Algnameabbr{} Outperforms Reasoning LLMs and Chain of Thought Prompting}
\label{sec:Reasoning}
We evaluate whether \Algnameabbr{} can match or surpass specialized reasoning methods. On the QASC dataset, general-purpose LLMs combined with \Algnameabbr{} outperform their reasoning-tuned counterparts, indicating that \Algnameabbr{} unlocks accuracy gains without dedicated reasoning fine-tuning. Furthermore, \Algnameabbr{} also surpasses chain-of-thought (CoT) prompting~\cite{wei2022chain}, which yields higher accuracy than CoT while avoiding the extra latency and token-generation overhead of producing reasoning traces. Both results are summarized in Table~\ref{tab:general_vs_reasoning_cot}, highlighting that \Algnameabbr{} offers an efficient, single-pass, and fine-tuning-free alternative to reasoning LLMs and CoT prompting on MCQA.

\begin{table}[t]
    \centering
    \resizebox{0.5\textwidth}{!}{
    \begin{tabular}{l|c}
        \toprule
         Model & Accuracy \\ 
         \midrule
         LLaMA-3-8B & 73.22\\
         DeepSeek-R1-Distill-Llama-8B & 73.33 \\
         LLaMA-3-8B with CoT & 74.62 \\
         LLaMA-3-8B with \Algnameabbr{} & \textbf{80.67}\\
         \midrule
         Qwen2-7B & 69.44 \\
         DeepSeek-R1-Distill-Qwen-7B & 70.63 \\
         Qwen2-7B with CoT & 74.19 \\
         Qwen2-7B with \Algnameabbr{} & \textbf{77.86} \\ \bottomrule
    \end{tabular}
    }
    \caption{General LLMs with \Algnameabbr{} outperform reasoning LLMs and chain of thought prompting on the QASC dataset.}
    \label{tab:general_vs_reasoning_cot}
\end{table}

\vspace{-1pt}
\subsection{Application to Quantized LLMs}
\vspace{-1pt}
To further explore \Algnameabbr{}'s application to memory‐constrained settings, we show its effectiveness on quantized LLMs, including LLaMA-3-8B, Mistral-7B-v0.1, and Qwen2-7B under 4-bit bitsandbytes (BNB) quantization.
As shown in Table~\ref{tab:booster}, \Algnameabbr{} effectively offsets accuracy loss caused by the aggressive compression, enabling each quantized model to outperform its full-precision version. 
This indicates \Algnameabbr{}'s potential for quantized LLMs in resource-constrained environments, without requiring additional training or external information.

\begin{table}[t]
    \centering
    \resizebox{0.5\textwidth}{!}{
    \begin{tabular}{l|ccc}
        \toprule
         Model & 4-bit & Full Precision & 4-bit + \Algnameabbr{} \\ 
         \midrule
         LLaMA-3-8B & 72.57 & 73.22 & \textbf{79.16} \\
         Mistral-7B-v0.1 & 57.45 & 57.78 & \textbf{65.23} \\
         Qwen2-7B & 67.71 & 69.44 & \textbf{78.83} \\ \bottomrule
    \end{tabular}
    }
    \caption{\Algnameabbr{} helps quantized LLMs outperform full-precision LLMs on the QASC dataset.}
    \label{tab:booster}
    \vspace{-2mm}
\end{table}

\begin{table}[t]
    \centering
    \resizebox{0.5\textwidth}{!}{
    \begin{tabular}{l|ccc}
        \toprule
         Model & w/o Attn Mask & w/o Pos Re-enc & \Algnameabbr{} \\ 
         \midrule
         LLaMA-3-8B & 19.22 & 67.39 & \textbf{80.67} \\
         Mistral-7B-v0.1 & 18.57 & 53.24 & \textbf{66.31} \\
         Qwen2-7B & 15.44 & 71.27 & \textbf{77.86} \\ \bottomrule
    \end{tabular}
    }
    \caption{Comparison of \Algnameabbr{} with that w/o attention masks or positional re-encoding.}
    \label{tab:albation}
    \vspace{-2mm}
\end{table}

\vspace{-1pt}
\subsection{Ablation Study}
\vspace{-1pt}
We demonstrate the individual contribution of attention masks and positional re-encoding to \Algnameabbr{}.
Specifically, we conduct experiments comparing \Algnameabbr{} with its ablated versions: without attention masks or positional re-encoding.
Experimental results for the LLaMA-3-8B, Mistral-7B-v0.1, and Qwen2-7B on the QASC dataset are given in Table~\ref{tab:albation}.
It is observed that each model has a significant loss of accuracy without attention masks or positional re-encoding, indicating that both attention structure and position encoding contributes to the self-ensemble process. 
\Algnameabbr{} has a special attention mask designed in Sections~\ref{sec:positional-encoding} and positional re-encoding in Section~\ref{sec:attention-mask} to enable intrinsic LLM inference over different choice groups and produce ensemble results efficiently.
This can effectively calibrate LLM confidence and enhance the accuracy on many-choice problems.

\subsection{Sensitivity to Group Size and Number of Trials}
\label{sec:sensitivity}
We study how \Algnameabbr{} depends on the group size $m$ and the number of ensembles $N$. Using Qwen2.5-7B on the QASC dataset, we sweep  $m \in \{3,4,5\}$ and $N \in \{10,20,40,80\}$. As shown in Table~\ref{tab:sensitivity-grid}, accuracy is relatively insensitive to the group size, while increasing the number of trials improves performance and stabilizes at $N \geq 40$, indicating robustness to hyperparameters. Consequently, \Algnameabbr{} can be used without hyperparameter tuning, reducing setup time while preserving performance.

\begin{table}[t]
\centering
\begin{tabular}{lccc}
\toprule
Trials $N$ & $m=3$ & $m=4$ & $m=5$ \\
\midrule
10 & 78.73 & 78.08 & 78.61 \\
20 & 79.59 & 80.67 & 79.91 \\
40 & 79.27 & 81.97 & 81.43 \\
80 & 80.45 & 82.40 & 81.97 \\
\bottomrule
\end{tabular}
\caption{Accuracy comparison of \Algnameabbr{} across group sizes and trial numbers.}
\label{tab:sensitivity-grid}
\end{table}

\section{Related Work}
\label{sec:related_work}
\paragraph{Confidence Calibration.}
Calibration aligns predicted probabilities with true correctness likelihoods to ensure reliable confidence estimates. Traditional methods like vector scaling~\cite{guo2017calibration}, isotonic regression~\cite{zadrozny2002transforming}, and Dirichlet calibration~\cite{kull2019beyond} have been adapted to LLMs but often assume fixed prediction heads and struggle with distribution shifts or multi-choice uncertainty. Recent generative-specific techniques, such as logit perturbation~\cite{jiang2021can}, prompt-based reweighting~\cite{zhou2022calibrating}, Bayesian post-hoc methods~\cite{wang2023posterior}, chain-of-though reasoning~\cite{sessler2024benchmarking} partially address these limitations. 
Moreover, pioneer work~\cite{zhong2025quantized} proposes a unified framework of calibrating LLMs under weight quantization, with theoretical foundations and practical design.
However, consistent calibration across diverse LLM architectures and tasks remains challenging, highlighting the need for effective, model-agnostic solutions.

\paragraph{LLM Ensemble.}
Ensembling has long been a powerful strategy to improve both model accuracy and robustness. In LLMs, methods such as voting schemes~\cite{zhou2022least}, stochastic decoding~\cite{ho2022large}, and diverse prompting~\cite{wang2022self} aggregate outputs from multiple model instances or inference runs. 
Notably, similar to our intuition of self-ensemble, a concurrent work~\cite{neeley2025survey} also proposes a "divide-and-conquer" strategy to improve LLM performance on healthcare tasks.
It has shown promising results in diagnosing rare diseases caused by genetic variants, a task that remains highly challenging.
These methods are especially useful for reducing variance in generative outputs and mitigating individual model biases. Additionally, ensemble approaches support calibration via confidence averaging and self-consistency mechanisms~\cite{wang2022self}. 
However, standard ensembles are computationally costly and less generalizable across model families and tasks, motivating for a more efficient and scalable ensemble method.






\section{Conclusion}
In this work, we demonstrate the confidence mis-calibration of LLMs generally on MCQA, particularly in the many-choice setting.
To solve this problem, we propose \Algnameabbr{} by decomposing a many-choice problem into several few-choice problems, and aggregating the intermediate results into the final solution. 
We further integrate \Algnameabbr{} with existing LLM architecture, enabling an intrinsic self-ensemble process for LLM inference.
Experimental results across various LLMs and datasets show that \Algnameabbr{} can effectively overcome this problem, improving confidence in correct answers and reducing confidence in incorrect answers.
This enables \Algnameabbr{} to significantly improve the accuracy of LLMs in MCQA by $8\%$ on average.

\section{Limitations and Potential Risks}
\label{sec:limitations}

In this work, we propose a \Algnameabbr{} to calibrate the LLM decisions on many-choice QA.
The application of \Algnameabbr{} is limited to the muti-choice question-answer~(MCQA) problems. 
While MCQA stands out as a standard and challenging benchmark to evaluate the ability of LLMs, there are open-ended QA tasks in real-world scenarios.
Calibrating LLMs' decisions on open-ended QA remains our future research.
Furthermore, \Algnameabbr{} is methodologically orthogonal to existing calibration techniques such as Vector Scaling and Dirichlet Calibration, enabling synergistic integration to further enhance performance.

\section*{Acknowledgments}
This research was supported by NSF CNS 2528780. We would like to thank The Center for Research Computing at Rice University for providing hardware and technical assistance for experiments. We extend special thanks to Xinyu Wang from the University of Kansas for help with the paper’s visualizations. The views and conclusions in this paper are those of the authors and do not represent the views of any funding or supporting agencies.

\bibliography{citation}

\clearpage
\appendix

\section*{Appendix}

\section{Efficiency Comparison of Multi-Pass and Single-Pass \Algnameabbr{}}

We follow existing work~\cite{wang2024taylor, liu2023winner, yuan2024kv} to benchmark the efficiency of multi-pass \Algnameabbr{} in Section~\ref{sec:self-ensumble} and the single-pass \Algnameabbr{} in Section~\ref{sec:plugin_self_ensemble} using LLaMA-3-8B on the QASC dataset. As shown in Table~\ref{tab:multi-vs-single}, single-pass \Algnameabbr{} successfully accelerates inference time without sacrificing accuracy while maintaining almost the same memory cost. This demonstrates the effectiveness of our Section~\ref{sec:plugin_self_ensemble} in accelerating the inference through parallelization.









\section{Confidence Mis-calibration Across Model Scales and Families}
We assess generality beyond 7B–8B models by evaluating LLaMA-3.2-1B/3B, LLaMA-3-8B, and Qwen2.5-1.5B/3B/7B/14B on QASC with $K\in\{2,4,6,8\}$. As shown in Table~\ref{tab:scale-mis-calibration}, accuracy consistently declines as $K$ increases for every model, indicating that the confidence mis-calibration phenomenon holds across architectures and scales.

\section{Extended Confidence Mis-calibration Evaluation}

To further illustrate \Algnameabbr{} mitigates confidence mis-calibration, we provide additional evaluations based on the under-confidence and over-confidence ratios. Here, under-confidence is defined as the proportion of correct answers where the model assigns probability below a threshold $\tau$, while over-confidence is the proportion of incorrect answers where the model assigns probability above $\tau$. For example, with $\tau=0.5$, the under-confidence ratio measures the percentage of correct answers where $f(Q)<0.5$, and the over-confidence ratio measures the percentage of incorrect answers where $f(Q)>0.5$.

We report results averaged across LLaMA-3-8B, Mistral-7B-v0.1, and Qwen2-7B on QASC, TruthfulQA, and MMLU-Pro. As shown in Table~\ref{tab:conf-ratio}, Self-Ensemble consistently reduces both under-confidence and over-confidence compared to standard inference, demonstrating its ability to mitigate confidence mis-calibration rather than exacerbate it.

\begin{table}[t]
\centering
\resizebox{0.5\textwidth}{!}{
\begin{tabular}{lcc}
\toprule
 & Multi-pass & Single-pass \\
\midrule
Time (s) per question $\downarrow$ & 0.78 & \textbf{0.16} \\
Memory (GB)  & 14.97 & 15.12 \\
Accuracy (\%)                            & 78.62 & 80.67 \\
\bottomrule
\end{tabular}
}
\caption{Efficiency and accuracy comparison of multi-pass vs. single-pass \Algnameabbr{}}
\label{tab:multi-vs-single}
\end{table}

\begin{table}[t]
\centering
\resizebox{0.5\textwidth}{!}{
\begin{tabular}{lcccc}
\toprule
Model & 2-choice & 4-choice & 6-choice & 8-choice \\
\midrule
LLaMA-3.2-1B   & 81.75 & 66.85 & 51.51 & 50.22 \\
LLaMA-3.2-3B   & 91.04 & 84.77 & 77.97 & 69.55 \\
LLaMA-3-8B     & 91.90 & 86.29 & 76.35 & 73.22 \\
Qwen2.5-1.5B   & 89.30 & 81.53 & 62.31 & 60.91 \\
Qwen2.5-3B     & 91.47 & 81.53 & 75.81 & 69.98 \\
Qwen2.5-7B     & 94.71 & 89.42 & 82.51 & 76.57 \\
Qwen2.5-14B    & 95.68 & 91.68 & 86.29 & 79.59 \\
\bottomrule
\end{tabular}
}
\caption{Accuracy of LLMs on the QASC dataset with different choice numbers.}
\label{tab:scale-mis-calibration}
\end{table}

\section{Additional \Algnameabbr{} Results Across Model Scales}
\label{app:scale-families}
To assess the generality of \Algnameabbr{}, we report 8-choice QASC accuracy across two model families spanning multiple scales. As shown in Table~\ref{tab:scale-combined}, \Algnameabbr{} consistently improves over standard inference from small (1--3B) to larger (7--14B) models.

\begin{table}[t]
\centering
\begin{tabular}{lcc}
\toprule
\multicolumn{3}{c}{Under-confidence $\downarrow$} \\
\midrule
Model & Standard & \Algnameabbr{} \\
\midrule
LLaMA-3-8B      & 43.45 & \textbf{35.33} \\
Mistral-7B-v0.1 & 58.33 & \textbf{50.60} \\
Qwen2-7B        & 41.41 & \textbf{34.25} \\
\midrule
\multicolumn{3}{c}{Over-confidence $\downarrow$} \\
\midrule
Model & Standard & \Algnameabbr{} \\
\midrule
LLaMA-3-8B      & 25.09 & \textbf{20.53} \\
Mistral-7B-v0.1 & 33.77 & \textbf{21.66} \\
Qwen2-7B        & 30.89 & \textbf{25.13} \\
\bottomrule
\end{tabular}
\caption{Under-confidence and over-confidence ratios at $\tau=0.5$.}
\label{tab:conf-ratio}
\end{table}

\section{Robustness to Random Grouping Seeds}
We quantify the effect of grouping randomness by repeating \Algnameabbr{} on QASC with seeds $s\in\{0,1,2,3,4\}$ that control the random partitioning of answer choices, holding all other variables fixed. Accuracy is reported as mean $\pm$ standard deviation across seeds. As shown in Table~\ref{tab:seed-robustness-table}, the small variances (all $\leq 0.60$ percentage points) indicate that \Algnameabbr{} is stable to grouping randomness across models.

\begin{table}[t]
\centering
\begin{tabular}{lcc}
\toprule
Model & Standard & \Algnameabbr{} \\
\midrule
\multicolumn{3}{c}{\textit{Qwen2.5 family}} \\
Qwen2.5-1.5B & 60.91 & \textbf{71.81} \\
Qwen2.5-3B   & 69.98 & \textbf{76.78} \\
Qwen2.5-7B   & 76.57 & \textbf{81.43} \\
Qwen2.5-14B  & 79.59 & \textbf{83.37} \\
\addlinespace
\multicolumn{3}{c}{\textit{LLaMA-3.x family}} \\
LLaMA-3.2-1B & 50.22 & \textbf{56.37} \\
LLaMA-3.2-3B & 69.55 & \textbf{77.32} \\
LLaMA-3-8B   & 73.22 & \textbf{80.67} \\
\bottomrule
\end{tabular}
\caption{\Algnameabbr{} consistently improves accuracy over standard inference across scales.}
\label{tab:scale-combined}
\end{table}

\section{Group-wise Normalization with a Null Option}
\label{app:group-normalization}
\Algnameabbr{} normalizes probabilities within each group by augmenting the group with a null choice ("None of the above") and applying a softmax over the augmented set. This guarantees per-group normalization (the probabilities over all candidates in a group sum to 1). Two immediate consequences explain the observed behavior: (i) when the correct answer is present, the null option receives negligible mass, so the sum over real options is close to 1; (ii) when the correct answer is absent, the null option absorbs probability mass, so the sum over real options drops below 1.

\begin{table}[t]
\centering
\begin{tabular}{lc}
\toprule
Model & Accuracy \\
\midrule
LLaMA-3-8B      & 80.24 $\pm$ 0.60 \\
Mistral-7B-v0.1 & 77.91 $\pm$ 0.60 \\
Qwen2-7B        & 65.87 $\pm$ 0.36 \\
\bottomrule
\end{tabular}
\caption{Accuracy with error bar of \Algnameabbr{} on QASC across random seeds.}
\label{tab:seed-robustness-table}
\end{table}

\section{Combination with Improvement}

\Algnameabbr{} is designed to be orthogonal to existing approaches and can be seamlessly combined with them to further enhance the performance and reliability.
Specifically, it can be integrated with efficient inference methods, such as KIVI~\cite{liu2024kivi, yuan2024kv} or H2O~\cite{zhang2023h2o}, to accelerate model execution while maintaining accuracy.
Moreover, it can also be combined with secure deployment techniques, such as Taylor-Unswift~\cite{wang2024taylor}, which effectively protects the intellectual property of LLMs through parameter space decomposition.
Finally, it is compatible with methods for multi-agent frameworks for solving complex tasks, such as ~\cite{chang2024main, yu2025survey}.

\section{Packages}
\label{sec:Packages}

In this work, we use the \texttt{transformers} along with \texttt{datasets} packages~\cite{wolf2019huggingface} for model and dataset loading.
All open-sourced packages have the Apache-2.0 license, which allows for academic research.
We use public benchmarks and open models strictly for research and evaluation in accordance with their original licenses/terms of use.
We release \Algnameabbr{} code, prompts, and evaluation scripts for research reproducibility; any derivatives of licensed datasets inherit their original research-only restrictions.

\begin{table}[t]
\centering
\begin{tabular}{l|c}
\toprule
Name & Value \\
\midrule
Data type & \texttt{torch.bfloat16} \\
Flash-Attention & False \\
Eval batch-size & 1 \\
Computing Infrastructure & GPU \\
GPU Model & NVIDIA-A40 \\
GPU Memory & 48GB \\ 
GPU Number & 1 \\
CUDA Version & 12.3 \\
CPU Memory & 512GB \\
\bottomrule
\end{tabular}
\caption{Experiment configuration and computing infrastructure.}
\label{tab:computing_infrastructure}
\end{table}
\section{Computational Infrastructure}
The computational infrastructure information is given in Table~\ref{tab:computing_infrastructure}.
\label{sec:hardware}

\end{document}